\documentclass{article}

\usepackage{PRIMEarxiv}

\usepackage[utf8]{inputenc} % allow utf-8 input
\usepackage[T1]{fontenc}    % use 8-bit T1 fonts
\usepackage{hyperref}       % hyperlinks
\usepackage{url}            % simple URL typesetting
\usepackage{booktabs}       % professional-quality tables
\usepackage{amsfonts}       % blackboard math symbols
\usepackage{nicefrac}       % compact symbols for 1/2, etc.
\usepackage{microtype}      % microtypography
\usepackage{lipsum}
\usepackage{fancyhdr}       % header
\usepackage{graphicx}       % graphics
\graphicspath{{media/}}     % organize your images and other figures under media/ folder

\usepackage{multirow}
\usepackage{amsmath}
\usepackage[linesnumbered,ruled,vlined]{algorithm2e}
\usepackage{pdfpages}
\usepackage{soul}
\usepackage{xcolor}

\title{Class-Specific Distribution Alignment for Semi-Supervised Medical Image Classification
}

\author{
  Zhongzheng Huang $^{1,2}$, Jiawei Wu $^{1,3}$, Tao Wang $^{1,4,}$\thanks{Corresponding authors.
\textit{\underline{Citation}}: 
\textbf{Huang, Z.; Wu, J.; Wang, T.; Li, Z.; Ioannou A. Class-Specific Distribution Alignment for Semi-Supervised Medical Image Classification. Computers in Biology and Medicine 2023, 164, 107280. https://doi.org/10.1016/j.compbiomed.2023.107280}}~~, Zuoyong Li $^{1,*}$ and Anastasia Ioannou $^{4,5}$ \\
  $^{1}$ College of Computer and Control Engineering, Minjiang University, China \\
  $^{2}$ College of Computer and Data Science, Fuzhou University, China \\
  $^{3}$ College of Electrical and Mechanical Engineering, Fujian Agriculture and Forestry University, China \\
  $^{4}$ International Digital Economy College, Minjiang University, China \\
  $^{5}$ Department of Computer Science and Engineering, European University Cyprus, Cyprus \\  
}

\begin{document}
\maketitle

\begin{abstract}
Despite the success of deep neural networks in medical image classification, the problem remains challenging as data annotation is time-consuming, and the class distribution is imbalanced due to the relative scarcity of diseases. To address this problem, we propose Class-Specific Distribution Alignment (CSDA), a semi-supervised learning framework based on self-training that is suitable to learn from highly imbalanced datasets. Specifically, we first provide a new perspective to distribution alignment by considering the process as a change of basis in the vector space spanned by marginal predictions, and then derive CSDA to capture class-dependent marginal predictions on both labeled and unlabeled data, in order to avoid the bias towards majority classes. Furthermore, we propose a Variable Condition Queue (VCQ) module to maintain a proportionately balanced number of unlabeled samples for each class. Experiments on three public datasets HAM10000, CheXpert and Kvasir show that our method provides competitive performance on semi-supervised skin disease, thoracic disease, and endoscopic image classification tasks.
\end{abstract}


\section{Introduction}\label{sec:introduction}
Computer vision and machine learning are now being extensively employed in the medical imaging domain for accurate and efficient disease diagnosis. Medical staff can choose computer-aided diagnosis as a viable companion tool to assist them in judging the type and severity of disease. Particularly, the renaissance of connectionism in computer vision has sparked interests in developing medical image analysis methods based on Convolutional Neural Networks (CNNs)~\cite{shen2017deep,MAO2022105729,esteva2017dermatologist}. CNNs offer a large learning capacity and typically perform favorably when a large amount of training data is available. However, due to the data bottlenecks and the expensive nature of clinician input, it is difficult for ordinary research institutions to collect and annotate large-scale medical image data~\cite{kaissis2020secure}. The rarity of some diseases also leads to the insufficiency of corresponding data~\cite{mascalzoni2014rare}, resulting in uneven class distributions of the datasets. Therefore, despite the remarkable progress being made, data-efficient learning in the medical imaging domain is still an important and challenging problem.

To alleviate the adverse effects caused by data and annotation scarcity, semi-supervised learning is widely used in medical image analysis. In broad terms, existing methods can be divided into generative model-based methods, graph-based methods, and consistency-based methods~\cite{yang2021survey}. Specifically, generative model-based methods~\cite{qi2018global, wu2019enhancing, liu2020regularizing} capture the data manifold by learning a generative model, and then generate novel training data to improve generalization. On the other hand, graph-based methods~\cite{kipf2016semi, wang2020graph} learn a graph embedding of the input dataset for propagating labels from labeled nodes to unlabeled nodes. In addition, consistency regularization methods~\cite{verma2019interpolation, li2020dividemix, sohn2020fixmatch} enforce the model to maintain a consistent output when different noises and perturbations are added to the data. However, all of these methods could be negatively impacted by class imbalance, which is especially common in the medical imaging domain. In particular, class imbalance will bias the model towards majority classes and away from minority classes, resulting in rapid performance degradation as the bias accumulates during training.

\begin{figure}[t]
\centering
\includegraphics[width=0.60\textwidth]{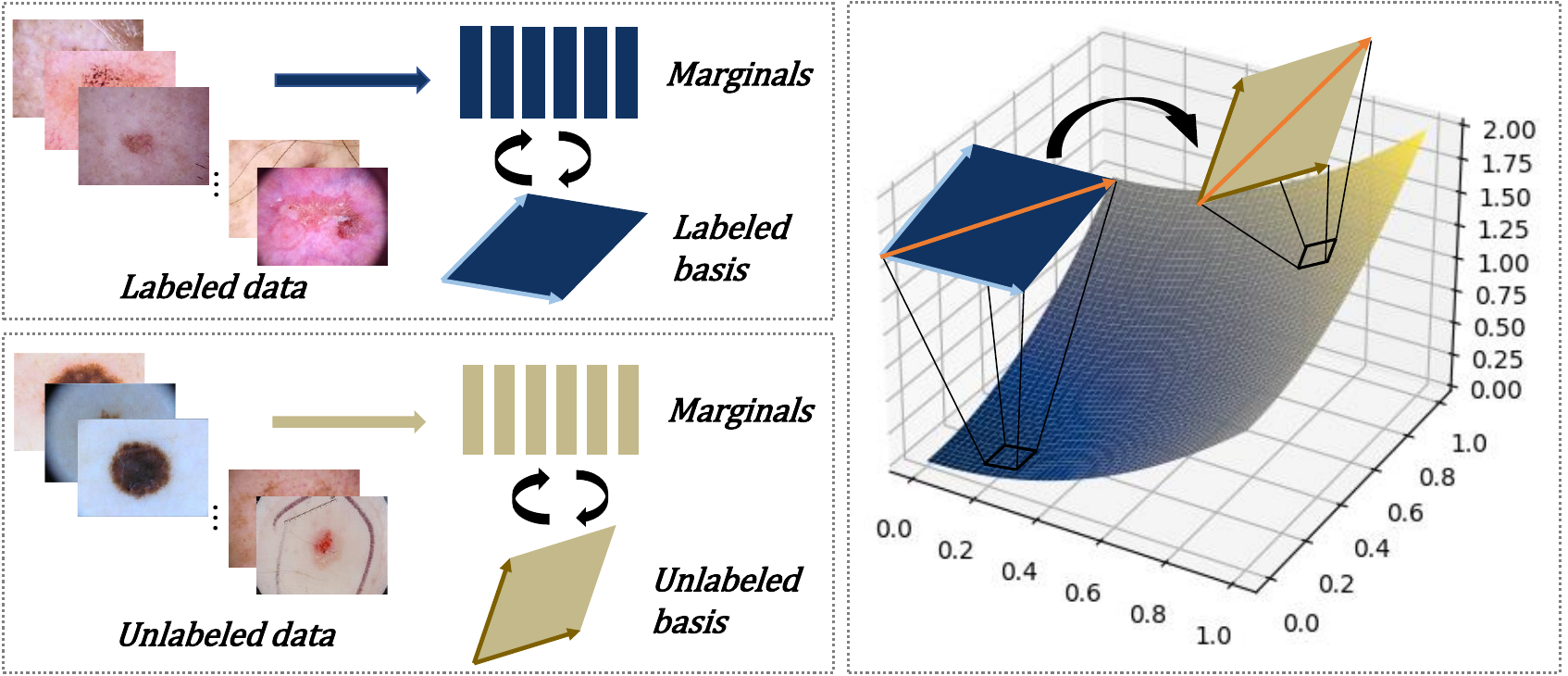}
\caption{Illustration of Class-Specific Distribution Alignment from a change of basis perspective. \textbf{Left: }The marginal prediction on labeled data and unlabeled data can be written with regard to their respective basis vectors. \textbf{Right: }Vectors in an $n$-dimensional vector space can be written in terms of both the labeled basis and the unlabeled basis, whereas a change of basis is required to minimize the distance between the two marginals.}
\label{fig:pathdemo1}
\end{figure}

\begin{figure*}[t!]
\centering
\includegraphics[width=0.9\textwidth]{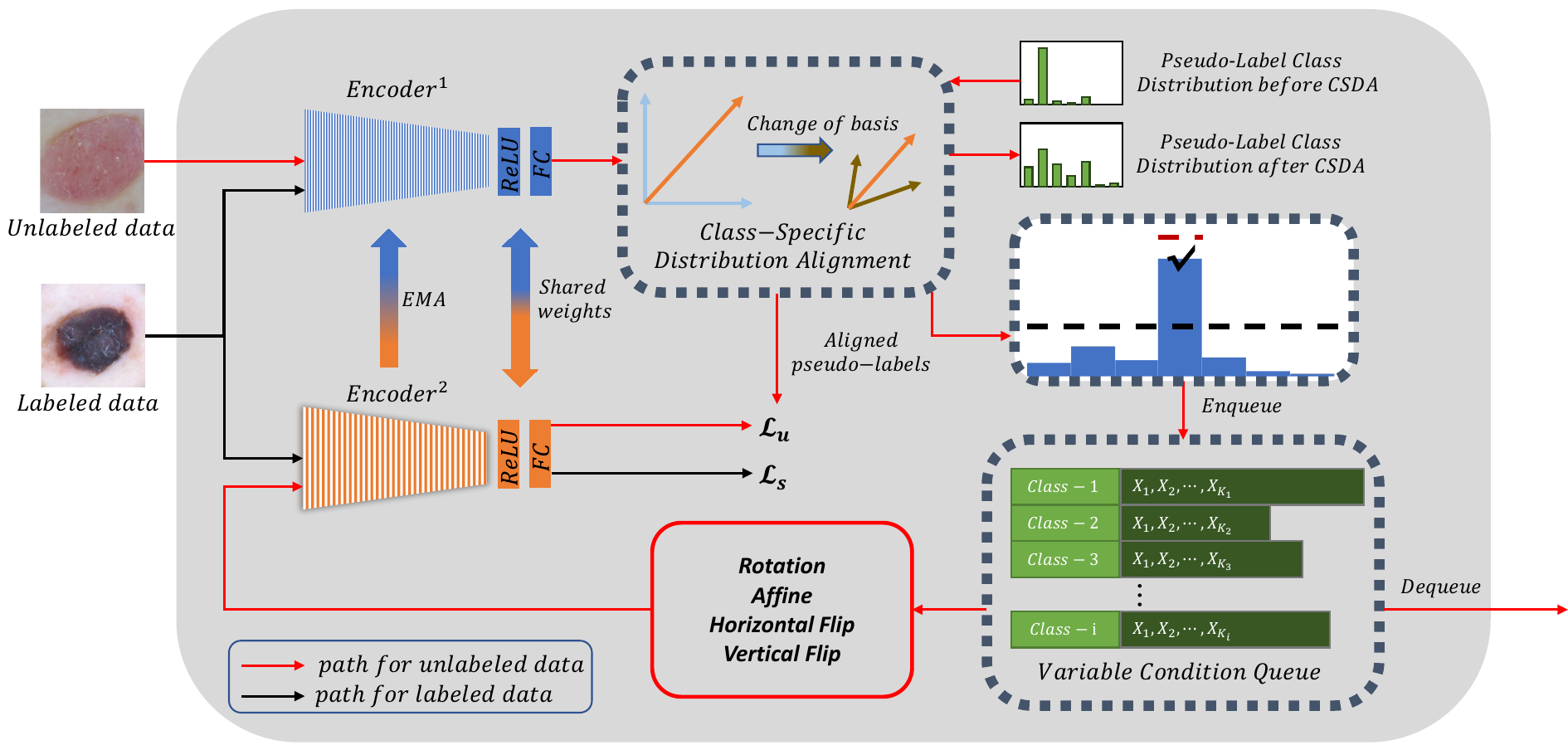}
\caption{Overview of the proposed learning framework. The model takes as input the labeled data and the unlabeled data at the same time. Predictions on the unlabeled data will be aligned by Class-Specific Distribution Alignment (Sec.~\ref{sec:csda} and Sec.~\ref{sec:csda4mic}), which avoids domination by majority classes and facilitates learning from minority classes by way of obtaining a more balanced pseudo-label class distribution. In addition, a Variable Condition Queue will maintain an appropriate number of unlabeled samples for each class (Sec.~\ref{sec:vcq}).}
\label{fig:pathdemo2}
\end{figure*}

To address this issue, Distribution Alignment (DA) first proposed by ReMixMatch~\cite{berthelot2019remixmatch} aims at aligning the marginal distribution of predictions on the labeled and unlabeled data, and demonstrates its effectiveness on unbalanced data in recent work such as  CReST~\cite{wei2021crest}. However, after exhaustive research, we observed that the theoretical foundation of DA is not well-studied.
In mathematics, elements in a vector space of finite dimension can be uniquely represented by coordinate vectors~\cite{anton2008elementary}. If two different bases are considered, the coordinate vector in one basis is usually different from the coordinate vector in the other basis. A change of basis consists of converting every assertion expressed in terms of coordinates relative to one basis into an assertion expressed in terms of coordinates relative to the other basis~\cite{kellaway1972linear}, as shown in Fig.~\ref{fig:pathdemo1}. In this work, we provide a novel perspective into DA by considering the process as a change of basis in the vector space spanned by marginal predictions.
In this way, we derive a more general formulation of Class-Specific Distribution Alignment (CSDA). In contrast to conventional DA, we consider class-disentangled marginal distributions to capture data distribution with a higher level of accuracy and to avoid the systematic bias resulting from class imbalance. Specifically, we estimate the labeled and the unlabeled marginal distribution and the mean class confidence of each category by exponential moving average. In addition, in the case of a highly skewed class distribution, we propose an update strategy to overcome the lack of update in minority classes. In particular, if a certain class is absent from the pseudo-labels at an iteration, we use an average scaling factor to estimate the marginal distribution and the mean class confidence for the unlabeled data from the labeled data of the same class. We also avoid the bias towards majority classes via class-specific temperature scaling.
The detailed update strategy is presented in Sec.~\ref{sec:csda4mic}.
Finally, we propose a Variable Condition Queue (VCQ) module in order to maintain a reasonable class distribution in every training batch for the unlabeled data. Specifically, VCQ implements a class-wise threshold for the maximum prediction probability to maintain queues of proportionately balanced unlabeled samples for model learning, improving the classification accuracy of minority classes. The overall framework of our method is summarized in Fig.~\ref{fig:pathdemo2}.

The main contribution of our work is three-fold:

\begin{itemize}
\item We provide a new perspective to Distribution Alignment via the change of basis technique, based on which we derive the Class-Specific Distribution Alignment to capture the correlation between marginal predictions across different categories. CSDA alleviates the negative impact of class imbalance in semi-supervised medical image classification, allowing us to obtain pseudo-labels for unlabeled data with better quality. 

\item We present a Variable Condition Queue module to maintain a proportionately balanced number of unlabeled samples for each class. The length of individual queues will change as the label confidence changes, in order to avoid under-sampling in minority categories and over-sampling in majority categories. 

\item We conduct extensive experiments on the publicly available dermatoscopic image dataset HAM10000, the chest X-ray dataset CheXpert and the endoscopic image dataset Kvasir. Results demonstrate that our method is superior or comparable to the state-of-the-art on the semi-supervised medical image classification tasks with all these datasets.
\end{itemize}

The rest of this paper is organized as follows. Sec.~\ref{sec:related} reviews the recent research on semi-supervised learning and the self-training framework, as well as skin disease, thoracic disease classification and endoscopic image classification. Sec.~\ref{sec:method} describes the CSDA framework and the VCQ module proposed in this paper and their technical details. Sec.~\ref{sec:experiments} reports and analyzes the experimental evaluation results. Finally, Sec.~\ref{sec:conclusion} concludes the paper with closing remarks.

\section{Related Work}
\label{sec:related}
In this section, we first review recent literature on semi-supervised learning to set the stage for our method. In particular, we focus on self-training methods that our work is based on, as well as methods that specifically address the class imbalance issue. Finally, we review methods proposed for medical image classification, i.e., skin lesion classification, thoracic disease classification, and endoscopic image classification, which are the three tasks used to validate the efficacy of our approach. 

\subsection{Semi-supervised Learning}
Semi-supervised learning uses a few labeled examples and a large amount of unlabeled data, from which the model will learn and generate more accurate predictions. As one of the most well-established areas of research in computer vision and machine learning, current focus have been to achieve effective learning with typically data-hungry deep models. Readers are referred to~\cite{yang2021survey} for a recent survey. In terms of methods developed for medical image classification, for example, Xie et al.~\cite{xie2021semi} design a confidence module according to class probability to improve performance based on the Mean-Teacher~\cite{tarvainen2017mean} framework. Bdair et al. \cite{bdair2021fedperl}~propose a method to encourage the unlabeled data to learn from peers to generate pseudo-labels for skin lesion classification. Guo et al.~\cite{guo2021semi} propose a data augmentation method according to the class activation map based on consistency regularization to improve the performance of interstitial lung disease classification. 

\noindent \textbf{Self-training.} The core idea of self-training is to use an existing classifier trained on the labeled data to obtain pseudo-labels for the unlabeled data and then retrain a new model with the pseudo-labels, which is a general semi-supervised learning method based on entropy minimization~\cite{feng2020dmt}. For instance, Lee et al.~\cite{lee2013pseudo} adopt online self-training to generate pseudo-labels after each forward pass in a network. Feng et al.~\cite{feng2020dmt} generate pseudo-labels through offline self-training based on the prediction between differently initialized models, then the model is fine-tuned with all labels. Mukherjee and Awadallah~\cite{mukherjee2020uncertainty} integrate Monte Carlo dropout into the self-training process through a Bayesian network and uses hard labels to calculate losses. Our work differs from the existing work in that we propose the Class-Specific Distribution Alignment framework and a Variable Condition Queue module to obtain pseudo-labels of higher quality when the class distribution is highly unbalanced.

\noindent \textbf{Unbalanced datasets.} In the case of an uneven class distribution, the aforementioned semi-supervised learning methods are prone to bias towards majority classes. Recently, some methods have been proposed to deal with this issue. For example, Fan et al.~\cite{fan2022cossl} propose a new co-learning framework to decouple representation and classifier learning. Kim et al.~\cite{kim2020distribution} propose a refinement method based on convex optimization to correct pseudo-labels. Wei et al.~\cite{wei2021crest} select more samples on minority categories according to an estimated class distribution; a progressive distribution alignment is combined into the sampling strategy. In contrast, we further decouple the data distribution into every class in order to reduce the bias resulting from class imbalance. In addition, we propose a variable condition queue to ensure that every class is proportionately represented in the unlabeled set used for training.

\subsection{Medical Image Classification}

In this paper, we validate the efficacy of our approach on three medical image classification tasks: skin lesion classification, thoracic disease classification, and endoscopic image classification. Therefore, we summarize recent literature in these areas as follows.

\noindent \textbf{Skin lesion classification.} Skin lesion classification based on dermoscopic imagery remains a challenging problem due to the large inter-class similarity and intra-class variation~\cite{liu2020semi, DONG2022106321,TALAVERAMARTINEZ2022105450}.
For instance, ARL~\cite{zhang2019attention} proposes an attention residual module that uses high-level feature maps to generate attention maps for lower levels to allow the model paying more attention to semantically important regions. In terms of weakly supervised learning methods, Shi et al.~\cite{shi2019active} adopt local sensitivity hashing and confidence ranking to filter unannotated data for active learning, and specifically propose an aggregation operation to improve classification performance. SRC~\cite{liu2020semi} designs the sample relation consistency paradigm based on Mean-Teacher to calculate the semantic relationship between different samples. GLM~\cite{gyawali2020semi} mixes the labeled data and the unlabeled data in the network input stage and the feature computation stage in order to regularize different portions of the network. NM~\cite{wang2021neighbor} designs a memory padding module to store the characteristics and the labels of labeled data, and a neighbor matching module to propagate the more accurate pseudo-labels. Unlike the above methods, our model specifically addresses the class imbalance issue by aligning class-specific data distributions and maintaining a queue of appropriately balanced unlabeled samples for semi-supervised learning.

\noindent \textbf{Thoracic disease classification.} Automatic disease classification by chest radiography has made great progress with the help of deep models~\cite{VOGADO2022105442}. For example, Ma et al.~\cite{ma2019multi} design a cross-attention network to obtain a more meaningful representation from data, and a multi-label balance loss function to address class imbalance. Sekuboyina et al.~\cite{sekuboyina2021relational} model the chest X-ray image as a multi-modal knowledge graph, and add additional nodes to improve performance by integrating auxiliary information. Sonsbeek et al.~\cite{sonsbeek2021variational} use electronic health records to help inference network based on knowledge distillation to learn the latent representation of X-ray images. Agu et al.~\cite{agu2021anaxnet} propose a model combining a detection module and an anatomical dependency module to classify chest diseases. The anatomical dependency module utilizes a graph convolution network to learn the relationship among the anatomical regions. Also considering the problems of insufficient data and unbalanced data distribution in chest X-ray classification, Sundaram and Hulkund~\cite{sundaram2021gan} use Generative Adversarial Networks (GANs) to generate data, which leads to higher downstream performance for the under-represented classes. Our contribution in this work, however, is orthogonal to the existing work as we propose to improve the self-training paradigm by learning distribution alignment in a class-specific fashion, and introduce several modules and techniques to address the class imbalance issue.

\noindent \textbf{Endoscopic image classification.} Deep learning has positively impacted the classification of endoscopic images~\cite{2020Kvasir}. For example, Srivastava et al.~\cite{srivastava2022video} propose a focal modulation network integrated with lightweight convolutional layers for better classifying small bowel anatomical landmarks and luminal findings. Shohei et al.~\cite{IGARASHI2020103950} develop a simple anatomical organ classifier which is effective in the data cleansing task for a collection of EGD images. Betul et al.~\cite{AY2022105725} design a reliable nasal aid system for identifying nasal polyps in endoscopic videos. Unlike existing methods in this area, our method focuses on the distribution characteristics of the data, and obtains a rebalanced class distribution on the unlabeled data to improve the classification performance.

\begin{table}[ht!]
\centering
\caption{Summary of the main notations.}
\setlength{\tabcolsep}{2mm}
\begin{tabular}{l|l}
\hline
Description                                        & Notation \\ \hline
Labeled dataset and unlabeled dataset              & $\mathcal{X},\mathcal{U}$    \\
Model prediction on $\mathcal{X}$ and $\mathcal{U}$ & $p(\cdot),q(\cdot)$    \\
Number of classes                                  & $n$    \\
Marginal prediction                                & $\boldsymbol{ \lambda^*_*}$ \\
Transition matrix for the change of basis              & $M$      \\
Mean class confidence                               & $c^*_*$   \\ \hline 
\end{tabular}
\label{tab:notations}
\end{table}

\section{Method}
\label{sec:method}

In this section, we first revisit distribution alignment from a change of basis perspective. Next, we propose a novel Class Specific Distribution Alignment (CSDA) method based on the change of basis that suits for medical image classification in class-imbalanced scenarios.
In addition, a Variable Condition Queue is proposed to further improve learning for minority classes via maintaining an appropriate number of unlabeled samples for each class.
\subsection{Class-Specific Distribution Alignment}
\label{sec:csda}

Let us begin with the problem at hand and the key notations. In semi-supervised learning, we are given a labeled training dataset $\mathcal{X}$ and an unlabeled training dataset $\mathcal{U}$. Here $\mathcal{X}=\{(x_k,y_k)\}_{k=1}^{K_X}$, where $x_k$ and $y_k$ are the $k$-th training sample and its corresponding class label, and $K_X$ is the total number of samples in the labeled training dataset. On the other hand, $\mathcal{U}=\{u_k\}_{k=1}^{K_U}$, where $u_k$ is the $k$-th training sample whose class label is not known, and $K_U$ is the total number of samples in the unlabeled training dataset. Given any test sample $u$, our goal is to predict $p_{\text{model}}(y|u;\theta)$, where $y$ is the prediction for the class label and $\theta$ stands for the model parameters. The main notations are summarized in Table~\ref{tab:notations}.

In this paper, we tackle the problem of empirical distribution mismatch, which is a key issue in semi-supervised learning (e.g., ~\cite{kamath2015learning, berthelot2019remixmatch, wang2019semi}). In principle, the labeled and the unlabeled training samples are assumed to be drawn from an identical distribution. However, due to the sampling bias, a considerable difference in the empirical distributions may appear, resulting in a degradation in the generalization abilities of trained models. ReMixMatch~\cite{berthelot2019remixmatch} offers a simple yet effective strategy known as Distribution Alignment (DA) to overcome this issue by aligning the model prediction on the unlabeled data $q(y|\mathcal{U})$ to the model prediction on the labeled data $p(y|\mathcal{X})$. In the following pages, let us revisit this approach from a change of basis perspective, which reveals a more general and flexible framework for DA. We begin by making two basic assumptions:

\textbf{Assumption 1.} We assume that the space of model prediction is $\mathcal{O} \in \mathbb{R}^n$, where $n$ is the number of classes. Denote the model prediction on the labeled data and the unlabeled data as $p(y|\mathcal{X})$ and $q(y|\mathcal{U})$, respectively. These two distributions are constrained with $p(y|\mathcal{X}) \cup q(y|\mathcal{U}) \subseteq \mathcal{O}$.

\textbf{Assumption 2.} Due to the sampling bias, there is a potential gap between $p(y|\mathcal{X})$ and $q(y|\mathcal{U})$. Consequently, the main goal of distribution alignment is to minimize the gap between $p(y|\mathcal{X})$ and $q(y|\mathcal{U})$.

As $p(y|\mathcal{X})$ and $q(y|\mathcal{U})$ are regarded as part of the same vector space $\mathcal{O}$, they can both be written with regard to their respective basis vectors, i.e., by writing them as a linear combination of class-dependent distributions. Let $ v_1^x, v_2^x, \dots, v_n^x$ and $ v_1^u, v_2^u, \dots, v_n^u$ be two bases of $\mathcal{O}$ for $p(y|\mathcal{X})$ and $q(y|\mathcal{U})$, respectively. We have:
\begin{equation}
\begin{split}
\label{eq:basis}
    p(y|\mathcal{X}) &= \begin{bmatrix} v_1^x & \dots & v_n^x \end{bmatrix} \begin{bmatrix} \boldsymbol{ \lambda^x_1} \\ \vdots \\ \boldsymbol{\lambda^x_n} \end{bmatrix} \\
    q(y|\mathcal{U}) &= \begin{bmatrix} v_1^u & \dots & v_n^u \end{bmatrix} \begin{bmatrix} \boldsymbol{ \lambda^u_1} \\ \vdots \\ \boldsymbol{\lambda^u_n} \end{bmatrix}
\end{split}
\end{equation}
where $\boldsymbol{ \lambda^x_i} \in \mathbb{R}^{1 \times n}$ and $\boldsymbol{ \lambda^u_i} \in \mathbb{R}^{1 \times n}, i=1,\dots,n$ are the marginal prediction of the $i$-th class on the labeled and the unlabeled data, respectively. For unlabeled data, the class membership is determined by the MAP estimate of the model prediction. We choose the MAP estimate as it is a simple and widely accepted means for image classification, and the mode of the prediction is somewhat robust to small perturbations. Therefore, Eqn.~\ref{eq:basis} provides a class-decoupled view to the model predictions.
Intuitively, distribution alignment seeks to achieve the following learning objective:
\begin{equation}
\label{eq:objective}
    \hat{\theta} = \arg\min_{\theta}D(p(y|\mathcal{X}),q(y|\mathcal{U}))
\end{equation}
where ${\theta}$ stands for the parameters of the neural network, and $D(\cdot,\cdot)$ is a distance function. Since it is unknown whether the bases for $p(y|\mathcal{X})$ and $q(y|\mathcal{U})$ are consistent, a change of basis is required to minimize Eqn.~\ref{eq:objective}. Assuming the transition matrix is $M \in \mathbb{R}^{n \times n}$, the relation between the bases can be written as follows: 
\begin{equation}
\label{eq3}
    [v_1^x \dots v_n^x]=[v_1^u \dots v_n^u] M^{-1}.
\end{equation} 
where $M$ is a nonsingular matrix:
\begin{equation}
\label{eq:transmatrix}
    M=\begin{bmatrix}
    m_{11} & \dots & m_{1n} \\
    \vdots & \ddots & \vdots \\
    m_{n1} & \dots & m_{nn}
    \end{bmatrix}
\end{equation}
Let $q(y|\mathcal{U}) = p(y|\mathcal{X})$, the coordinates transformation corresponding to Eqn.~\ref{eq3} can be written as:
\begin{equation}
    [v_1^u \dots v_n^u] M^{-1} \begin{bmatrix} \boldsymbol{\lambda^x_1} \\ \vdots \\ \boldsymbol{\lambda^x_n} \end{bmatrix} =
    [v_1^u \dots v_n^u] \begin{bmatrix} \boldsymbol{\lambda^u_1} \\ \vdots \\ \boldsymbol{\lambda^u_n} \end{bmatrix}
\end{equation}
\begin{equation}
\label{eq4}
    \begin{bmatrix} \boldsymbol{\lambda^x_1} \\ \vdots \\ \boldsymbol{\lambda^x_n} \end{bmatrix} =
    M \begin{bmatrix} \boldsymbol{\lambda^u_1} \\ \vdots \\ \boldsymbol{\lambda^u_n} \end{bmatrix} = 
    \begin{bmatrix}
    m_{11} & \dots & m_{1n} \\
    \vdots & \ddots & \vdots \\
    m_{n1} & \dots & m_{nn}
    \end{bmatrix}
    \begin{bmatrix} \boldsymbol{\lambda^u_1} \\ \vdots \\ \boldsymbol{\lambda^u_n} \end{bmatrix}
\end{equation}
\noindent where
\begin{equation}
    \label{eq5}
    \boldsymbol{\lambda_i^x}\big\vert_{j} =
    \text{Normalize}(\sum_{k=1}^{n} m_{ik}\boldsymbol{\lambda^u_k}\big\vert_{j})
\end{equation}
\noindent where $\boldsymbol{\lambda_i^x}\big\vert_{j}$ denotes the $j$-th element of $\boldsymbol{\lambda_i^x}$, $\boldsymbol{\lambda^u_k}\big\vert_{j}$ denotes the $j$-th element of $\boldsymbol{\lambda^u_k}$ , and the added $\text{Normalize}(\cdot)$ operation ensures that $\boldsymbol{\lambda_i^x}$ is a valid distribution after the transformation.

Eqn.~\ref{eq5} is a rather general formulation for distribution alignment, as we have established pairwise relations between the marginal prediction of a given class $i$ on the labeled data $\boldsymbol{\lambda^x_i}$ and the marginal prediction of all classes on the unlabeled data $\boldsymbol{\lambda^u_k}, k=1,\dots,n$. Although it is possible to consider this general case, by setting $M$ to a diagonal matrix (i.e., $m_{ij}=0, \forall i \neq j$) we arrive at a more familiar version of Eqn.~\ref{eq5}:
\begin{equation}
    \label{eq51}
    \boldsymbol{\lambda_i^x} =
    \text{Normalize}(m_{ii}\boldsymbol{\lambda^u_i})
\end{equation}

\noindent Since $\boldsymbol{\lambda^x_i}$ and $\boldsymbol{\lambda^u_i}$ are both $n$-dimensional distributions, we could convert $m_{ii}$ into a vector and perform distribution alignment by:
\begin{equation}
\label{eq6}
\boldsymbol{m_{ii}} = \boldsymbol{\lambda_i^x} \oslash \boldsymbol{\lambda_i^u}, \boldsymbol{m_{ii}} \in \mathbb{R}^{1 \times n}
\end{equation}
where $\oslash$ denotes the Hadamard division. In this way, $\boldsymbol{m_{ii}}$ becomes a measure to quantify the class-wise distance between two distributions.
Given the model's prediction $q = p_{\text{model}}(y|u;\theta)$ on an unlabeled sample $u$, the definition of Class-Specific Distribution Alignment (CSDA) can be written as:
\begin{equation}
    \tilde{q} = \text{Normalize}(q \otimes \boldsymbol{\lambda_i^x} \oslash \boldsymbol{\lambda_i^u})
    \label{eq:cdda}
\end{equation}
where $\tilde{q}$ is the aligned label guess for $u$ and $\otimes$ denotes the Hadamard product. The class membership $i$ for the unlabeled sample $u$ is determined by the MAP estimate of $q$ (i.e., $i=\arg\max_{y}q=\arg\max_{y}p_{\text{model}}(y|u;\theta))$. It is now clear that the vanilla Distribution Alignment (DA) introduced in ReMixMatch can be seen as a special case of the above, where there is a lack of subscript for marginal distributions $\boldsymbol{\lambda^x}$ and $\boldsymbol{\lambda^u}$. This suggests that in vanilla DA we do not consider the class-dependent distributions, but to only keep class-agnostic marginal distributions.

\subsection{Class-Specific Distribution Alignment for Semi-supervised Medical Image Classification}
\label{sec:csda4mic}
The CSDA formulation in Eqn.~\ref{eq:cdda} provides a general avenue to aligning class-wise marginal distributions on labeled data and unlabeled data. In this section, let us move on to discuss practical considerations in implementing CSDA for semi-supervised medical image classification. In addition to the conventional challenges for semi-supervised learning such as the scarcity of labeled data, the class distribution in medical images could also be highly imbalanced as some diseases or conditions can be rare by nature. Therefore, we use mean class confidence as an additional proxy to deal with the class imbalance problem.

Following~\cite{berthelot2019remixmatch}, we compute the labeled set distribution $\boldsymbol{\lambda^x_i}$ of each class by Exponential Moving Average (EMA). The formulation for EMA is typically written in a recursive form as follows:
\begin{equation}
    \text{EMA}_t = \text{EMA}_{t-1} * \omega  + z_t*(1 - \omega)
    \label{eqn:ema}
\end{equation}

\noindent where $\text{EMA}_t$ is the EMA value for the current step $t$. $z_t$ is the data value being analyzed for the current time step $t$, such as the probability values in the labeled set distribution. $\text{EMA}_{t-1}$ is the EMA value for the previous time step $t-1$. $\omega$ ($0 < \omega < 1$) is a momentum coefficient which is a hyperparameter that we set empirically to 0.95 in all our experiments to allow a stable update for the marginal distributions. It should be noted that using EMA for both labeled and unlabeled distributions allow us to perform distribution alignment with the current state of network parameters in a dynamic fashion as the training progresses, and the update in Eqn.~\ref{eqn:ema} is simple and efficient. In addition, we also use EMA to compute the mean class confidence $c^x_i$ of each class. For a given minibatch of training data, the mean class confidence for the $i$-th class can be written as:
\begin{equation}
    c^x_i = \frac{1}{K_i}\sum_{k=1}^K \big(\boldsymbol{\lambda_i^{x_k}} \big\vert_{i}\big)
\end{equation}

\noindent where $\boldsymbol{\lambda_i^{x_k}}$ is the model prediction for the $k$-th image and $K_i$ is the number of labeled images in the minibatch that belong to the $i$-th class. $\boldsymbol{\lambda_i^{x_k}} \big\vert_{i}$ denotes the $i$-th element of $\boldsymbol{\lambda_i^{x_k}}$. Essentially, $c^x_i$ is the mean of the model prediction for the $i$-th class for all images that belong to the $i$-th class in the minibatch.
The unlabeled set distribution $\boldsymbol{\lambda^u_i}$ and the corresponding mean class confidence $c^u_i$ are also computed with EMA in a similar fashion. Again, for unlabeled data, the class membership is determined by the MAP estimate of the model prediction.

In the case of a highly skewed class distribution, the model prediction will be biased and the EMA update of the unlabeled marginal prediction and the mean confidence for minority classes will become infrequent, affecting the ability of the model in classifying these classes. As a result, we propose the following update equation for $c_i^u$:
\begin{equation}
    c_i^{u,(t)}=\begin{cases}
        c_i^x*\frac{1}{n}\sum\limits_{i=1}^{n} c_i^{u,(t-1)}/c_i^x, & c_i^u = 0 \\
        c_i^{u,(t-1)}*\omega+c_i^u*(1-\omega), & c_i^u \neq 0 \\
    \end{cases}\label{update_unlabel}
\end{equation}

\noindent where $c_i^u$ is the unlabeled mean class confidence in the current batch, $c_i^{u,(t-1)}$ and $c_i^{u,(t)}$ are the moving average of $c_i^u$ at time step $t-1$ and $t$, respectively. Eqn.~\ref{update_unlabel} states that if the $i$-th class is never the mode of prediction for any unlabeled image in a certain batch (i.e., $c_i^u = 0$), we replace it with a reasonable estimate $c_i^x*\frac{1}{n}\sum\limits_{i=1}^{n} c_i^{u,(t-1)}/c_i^x $, in which $\frac{1}{n}\sum\limits_{i=1}^{n} c_i^{u,(t-1)}/c_i^x $ is an average scaling factor that converts $c_i^x$ into $c_i^u$. Otherwise, we use EMA with momentum coefficient $\omega$ to update $c_i^{u,(t)}$. The rest of the unlabeled marginal distribution is also updated in this way.

Furthermore, due to data scarcity and class imbalance, we also observe that $\boldsymbol{\lambda_i^x}$ can be biased towards the majority classes over update iterations. Therefore, we use temperature scaling that impels the labeled set distribution in a certain state close to the ground truth with reference to the Boltzmann distribution. The temperature for the $i$-th class is defined as $T_i$($0 < T_i < 1$), and we
use $\text{Normalize}\Big(\big(\boldsymbol{\lambda_i^x}\big)^{T_i}\Big)$ rather than $\boldsymbol{\lambda_i^x}$ as the labeled set distribution. In this way, as the temperature decreases, the distribution after scaling is closer to uniform so that the bias towards majority classes could be alleviated. As $T_i$ reaches a suitable value, the distribution will gradually approach thermodynamic equilibrium, leading to more balanced model predictions. In this work, we need to consider class-specific parameters to control the rebalancing strength for each class. A natural choice, therefore, would be the mean class confidence $c_i^x$, as we empirically found that it is positively related to the number of labeled data in each class. So we set $T_i$ to be negatively related to $c_i^x$ as follows:
\begin{equation}
    T_{i} = 1 - c_i^x
    \label{eqn:temperature}
\end{equation}

\noindent Such a choice of temperature scaling makes $T_i$ lower for majority classes, hence the prediction bias towards majority classes can be prevented. When $T_{i} \to 1$, $\boldsymbol{\lambda_i^x}$ will be transformed mildly. On the other hand, when $T_{i} \to 0$, $\boldsymbol{\lambda_i^x}$ will be smoothed out to prevent the prediction bias towards majority classes. It should be noted that, it is possible to use other functional forms in Eqn.~\ref{eqn:temperature}, but we choose the current relationship as it is one of the simplest forms of class-specific temperature scaling. In particular, we compared our scaling function with constant temperatures to show its superiority, and the results are presented in Fig.~\ref{fig5} (b) and (c).

\subsection{Variable Condition Queue}
\label{sec:vcq}

Next, let us discuss the proposed variable condition queue for self-training. In this work, we adopt the self-training strategy~\cite{grandvalet2004semi} for annotation efficient learning from partially labeled medical images. Self-training is an iterative algorithm that trains a classifier with labeled data at first, and then uses this classifier to classify unlabeled data and select the class with highest confidence as pseudo-labels. More specifically, given an unlabeled sample $u$, we first obtain its aligned label guess $\tilde{q}$ with Eqn.~\ref{eq:cdda} and then use the MAP estimate of $\tilde{q}$ (i.e., $\arg\max_{y}\tilde{q}$, $y \in \mathcal{Y}$, where $\mathcal{Y}$ is the label space) as the pseudo-label for $u$. Conventionally, self-training combines the unlabeled samples and their generated pseudo-labels with the labeled data to train the classifier for the next iterative step.

The unlabeled set $\mathcal{U}$ plays an important role in updating the parameters of the model, particularly in the self-training framework. As mentioned above, the pseudo-labels are used together with truth labels. However, errors from pseudo-labels could negatively impact model updates, especially early on in the training process, causing a significant decrease in the model performance. Another important consideration is the additional computational budget required for training with unlabeled data. For example, the usual training process back-propagates the gradients of all the unlabeled samples, which could lead to an unmanageable amount of computation time and a highly unbalanced class distribution.

To address the above issue, we propose a variable condition queue module to keep unlabeled samples in a class-adaptive manner with threshold filtering. Specifically, we first set the maximum number of samples (queue length) for the $i$-th class to $\text{len}_i$ as follows:
\begin{equation}
    \text{len}_{i} = \lfloor L*\frac{(c_i^x)^\gamma}{\sum_{i=1}^n (c_i^x)^\gamma} \rfloor
    \label{queuelength}
\end{equation}
where $L$ is the maximum length of the queue and $\gamma$ is a resampling parameter. There are some typical choices for $\gamma$ in the existing research~\cite{galdran2021balanced}. Choosing $\gamma = 1$ amounts to selecting class samples where the length is similar to the class frequency in the training set, i.e., \textit{instance-based sampling}, while choosing $\gamma = 0$ results in a uniform queue length $\text{len}_{i} = \lfloor L / n \rfloor$ of all classes, which is called \textit{class-based sampling}. Another choice is the \textit{square-root sampling} that chooses $\gamma = 1 / 2$ \cite{galdran2021balanced}, resulting in a middle ground between the two sampling methods above. Additionally, we set a class-wise minimum threshold $\tau_i$ for the maximum prediction probability $\max(\tilde{q})$:
\begin{equation}
    \tau_{i}=\min(c_i^u, \delta)
    \label{condition_threshold}
\end{equation}
where $\tau_{i}$ is the threshold for minimum class probability, $\delta$ is a hyper-parameter for the minimum prediction probability to enqueue. To sum up, given an unlabeled sample and its aligned label guess pair $(u, \tilde{q})$, our condition to enqueue can be specified as:
\begin{equation}
    \text{Enqueue}(u, \tilde{q}) \Longleftrightarrow [\arg\max_y \tilde{q}=i] \cap [\max(\tilde{q})>\tau_i]
    \label{condition_enqueue}
\end{equation}
\noindent where $\text{Enqueue}(\cdot)$ denotes the enqueue operation.

We note that when there are samples in the queue, standard data augmentation will also be applied to the unlabeled samples so as to prevent slow queue updates. We use this queue of unlabeled data to replace the original unlabeled dataset for loss computation to avoid the accumulation of labeling errors and to reduce computational complexity. In addition, by setting an appropriate queue length for each class we could learn to classify minority classes with better quality.

\subsection{Model Learning}
\label{sec:loss}
In this work, our learning objective is to minimize the loss function $\mathcal{L}$ on both labeled data and unlabeled data, following the common practice in semi-supervised learning:
\begin{equation}
    \mathcal{L} = \mathcal{L}_{s} + \eta \mathcal{L}_{u}
\end{equation}
where $\mathcal{L}_{s}$ is the loss function on the labeled data based on their truth labels, $\mathcal{L}_u$ is the loss function on the unlabeled data in the variable condition queue based on their pseudo-labels, and $\eta$ is the relative weight between the two terms. We use cross-entropy loss for both $\mathcal{L}_{s}$ and $\mathcal{L}_{u}$, noting that $\mathcal{L}_{u}$ is calculated using soft labels that encode the probability distribution across different classes. In our implementation, we follow~\cite{berthelot2019mixmatch} and~\cite{tarvainen2017mean} to linearly ramp up the weight of the unsupervised loss term from 0 to its final value, which is set to 1 throughout all our experiments. More specifically, the value of $\eta$ is given as:
\begin{equation}
    \eta = \text{epochs}_{t} ⁄ \text{epochs}
\end{equation}
\noindent where $\text{epochs}_{t}$ and $\text{epochs}$ are the current training epoch and the total number of training epochs, respectively. The full algorithm for CSDA is summarized in Algorithm \ref{alg:algorithm1}, where $H(\cdot, \cdot)$ denotes the cross entropy loss.

In addition, we adopt a two-stream iterative approach~\cite{wang2021neighbor} to train the feature encoder of our model, as illustrated in Fig.~\ref{fig:pathdemo2}. Specifically, we keep two separate sets of weights $\theta_e^1$ and $\theta_e^2$ for the feature encoder while sharing a common set of subsequent fully connected layers $\theta_{\text{fc}}$. During training (1) we first train $\theta_e^1$ and $\theta_{\text{fc}}$ via back-propagation with $\mathcal{L}_s$, using labeled data only for one epoch, then apply CSDA to obtain pseudo-labels for unlabeled data and use VCQ to initialize the unlabeled data queue. (2) Then, we train $\theta_e^2$ and $\theta_{\text{fc}}$ via back-propagation with $\mathcal{L}_s$ and $\mathcal{L}_u$, using both the labeled data and the unlabeled data queue for another epoch. (3) This is followed by updating $\theta_e^1$ with EMA such that $\theta_e^{1,(t)} = \theta_e^{1,(t-1)} * \omega + \theta_e^{2,(t-1)} * (1 - \omega) $, where $t$ is the current time step. (4) We then apply CSDA and VCQ again, in preparation for training $\theta_e^2$ and $\theta_{\text{fc}}$ for another epoch. Steps (2) to (4) above are repeated for a fixed number of times to obtain our final feature encoder $\theta_e^{2}$. We note that the delayed update of $\theta_e^1$ through EMA improves the stability of the model parameters. See Sec.~\ref{sec:implementation} for details.

\begin{algorithm}
\caption{Class-Specific Distribution Alignment.}\label{alg:algorithm1}
\textbf{Input:} Labeled training dataset $\mathcal{X}$, unlabeled training dataset $\mathcal{U}$, max epochs $N$ \;
\textbf{Output:} Parameters of the network $\theta_e^1$, $\theta_e^2$, $\theta_{\text{fc}}$ \;
Initialize $\theta_e^1$ and $\theta_{\text{fc}}$ by training on $\mathcal{X}$ for one epoch
Obtain pseudo-labels by CSDA and use VCQ to initialize the unlabeled data queue \;
\For {$epoch = 2$ to $N$}
{
    Sample batches of $\mathcal{X}$ and $\mathcal{U}$ as the input to $\theta_e^2$ \;
    Forward pass to compute the supervised loss: $\mathcal{L}_{s} = \frac{1}{K} \sum_{k=1 \dots K} H(\theta_{\text{fc}}(\theta_e^2(x_{k})), y_{k})$ \;
    Forward pass to compute the unsupervised loss: $\mathcal{L}_{u} = \frac{1}{L} \sum_{l=1 \dots L} H(\theta_{\text{fc}}(\theta_e^2(u_{k})), \tilde{y_{k}})$ \;
    Backward pass to update the parameters of the encoder $\theta_e^2$ and the fully connected layers $\theta_{\text{fc}}$ \;
    Update parameters of $\theta_e^1$ by EMA \;
    Update labeled and unlabeled set distribution ${\boldsymbol{\lambda^{x}_*}}$, ${\boldsymbol{\lambda^{u}_*}}$, labeled set confidence $c^x_*$ by EMA, unlabeled set confidence $c^u_*$ by Eqn.~\ref{update_unlabel}, adjust ${\boldsymbol{\lambda^{x}_*}}$ by temperature scaling with Eqn.~\ref{eqn:temperature} \;
    Align pseudo-labels $q$ by CSDA with Eqn.\ref{eq:cdda} \;
    Update VCQ according to Eqn.~\ref{queuelength} $\sim$ Eqn.~\ref{condition_enqueue} \;
}
\end{algorithm}

\section{Experiments}
\label{sec:experiments}
We evaluate our method on the publicly available dermoscopic image classification dataset HAM10000~\cite{tschandl2018ham10000, codella2019skin}, thoracic disease classification dataset CheXpert~\cite{irvin2019chexpert} and endoscopic image classification dataset Kvasir~\cite{2020Kvasir}. Our method provides competitive performance on all datasets, and we also present ablation experiments to validate the efficacy of individual components of our approach.

\subsection{Setup}

\subsubsection{Skin disease classification}
The HAM10000 dataset has a total of 10,015 high quality dermoscopic images with reliable diagnoses, composed of 7 types of skin diseases: melanoma (MEL), melanocytic nevi (NV), Basal cell carcinoma (BCC), actinic keratoses (AKIEC), benign keratosis-like lesions (BKL), dermatofibroma (DF), and vascular lesions (VASC). The class distribution is highly imbalanced; for example, NV accounts for about 67 percent of the total number of images. We adopt AUC (area under the ROC curve) and MCA (mean class accuracy) as our evaluation metrics. We follow the dataset split in~\cite{wang2021neighbor} for the training, validation and test sets.

\subsubsection{Thoracic disease classification}
CheXpert is a publicly available dataset of 224,316 frontal and lateral chest radiographs of 65,240 patients. The dataset divides chest radiographic observations into 14 categories, 12 of which indicate whether the patients have 12 disease characteristics such as cardiac hypertrophy and lung lesions. Since it is impossible to obtain completely correct diagnostic results from chest radiographs, the output of each category includes three options: positive, negative and uncertain. In this paper, we follow the experiment settings in~\cite{wang2021neighbor} by removing all uncertain and lateral-view samples from the dataset and use the public data split from~\cite{gyawali2019semi}.

\subsubsection{Endoscopic image classification}
Kvasir is a large Video Capsule Endoscopy (VCE) dataset collected from examinations at Hospitals in Norway. It has annotated and medically verified 47,238 frames with a bounding box around detected anomalies from 14 different classes of findings. We randomly select 70$\%$ of images from this dataset as the training set, 10$\%$ as the validation set, and the remaining 20$\%$ as the test set.

\subsubsection{Implementation Details}
\label{sec:implementation}
For HAM10000, we follow~\cite{wang2021neighbor} and randomly sampled each category of the dataset and selected 50 and 10 samples as the test and validation sets, respectively, and treat the remaining samples as the training set, dividing it into labeled and unlabeled sets. The proportion of labeled data in the training set is 1.8$\%$, 3.6$\%$, 6$\%$, 8$\%$, and 12$\%$, respectively (i.e., 175, 350, 600, 800 and 1200 labeled samples). Note that with 175 labeled images, there is only one shot in the minority classes. Each image is rotated within $\pm$ 10 degrees, resized to 128 $\times$ 128 to match the input requirements of the network, and translated in the range of (0.1, 0.1) fraction of the image. For CheXpert, we apply the same data augmentation, and use the data splits from~\cite{gyawali2019semi}.

We implement our method with PyTorch~\cite{paszke2019pytorch} and use a single NVIDIA GeForce RTX 3090 GPU for training the neural network. For a fair comparison, we follow~\cite{gyawali2020semi} and~\cite{wang2021neighbor} to use AlexNet~\cite{krizhevsky2012imagenet} as the backbone of our model and the network output layer of~\cite{gyawali2020semi}.
For this model, the FLOPs are 47.933M and the number of parameters is 9.692M.
The number of training epochs is set to 256, and the batch size is set to 128 (for labeled data). The mini-batch SGD algorithm is used for network optimization with an initial learning rate of $1e^{-4}$. A multi-step decay is also used for the learning rate, by multiplying it with 0.1 at the 50th and 125th epoch, respectively.

\begin{table*}[t!]
\centering
\caption{Performance on the HAM10000 dataset. The first row indicates the number of labeled training samples.}
\label{tab1}
\setlength{\tabcolsep}{2.5mm}
\begin{tabular}{l|cc|cc|cc|cc|cc}
\hline
\multirow{2}{*}{Method}             & \multicolumn{2}{c|}{175}                                  & \multicolumn{2}{c|}{350}                                  & \multicolumn{2}{c|}{600}                                  & \multicolumn{2}{c|}{800}                                  & \multicolumn{2}{c}{1200}                                  \\ \cline{2-11} 
                                    & \multicolumn{1}{c|}{AUC}             & MCA                & \multicolumn{1}{c|}{AUC}             & MCA                & \multicolumn{1}{c|}{AUC}             & MCA                & \multicolumn{1}{c|}{AUC}             & MCA                & \multicolumn{1}{c|}{AUC}             & MCA                \\ \hline
Baseline                            & \multicolumn{1}{c|}{0.6880}          & 0.2865             & \multicolumn{1}{c|}{0.7545}          & 0.3980             & \multicolumn{1}{c|}{0.7689}          & 0.3665             & \multicolumn{1}{c|}{0.8098}          & 0.4187             & \multicolumn{1}{c|}{0.8311}          & 0.4269             \\ 
FixMatch\cite{sohn2020fixmatch} & \multicolumn{1}{c|}{0.7028}          & 0.2923             & \multicolumn{1}{c|}{0.7564}          & 0.3668             & \multicolumn{1}{c|}{0.7747}          & 0.3954             & \multicolumn{1}{c|}{0.8002}          & 0.4384             & \multicolumn{1}{c|}{0.8197}          & 0.4298             \\ 
Self-train\cite{grandvalet2004semi}                          & \multicolumn{1}{c|}{0.7008}          & 0.2981             & \multicolumn{1}{c|}{0.7608}          & 0.3952             & \multicolumn{1}{c|}{0.7691}          & 0.3967             & \multicolumn{1}{c|}{0.8108}          & 0.4127             & \multicolumn{1}{c|}{0.8350}          & 0.4783             \\ 
GLM\cite{gyawali2020semi}      & \multicolumn{1}{c|}{\textbf{0.7214}}          & 0.2987          & \multicolumn{1}{c|}{0.7764}          & 0.3876          & \multicolumn{1}{c|}{0.7831}          & 0.4229          & \multicolumn{1}{c|}{0.8112}          & 0.4474          & \multicolumn{1}{c|}{0.8270}          & 0.5012          \\ 
NM\cite{wang2021neighbor}        & \multicolumn{1}{c|}{0.7051}          & 0.3069             & \multicolumn{1}{c|}{0.7731}          & 0.4180             & \multicolumn{1}{c|}{0.7918}          & 0.4217             & \multicolumn{1}{c|}{0.8220}          & 0.4501             & \multicolumn{1}{c|}{0.8399}          & 0.4729             \\
BiS\cite{2021Rethinking}        & \multicolumn{1}{c|}{0.6919}          & 0.2866             & \multicolumn{1}{c|}{0.7598}          & 0.3859             & \multicolumn{1}{c|}{0.7699}          & 0.3088             & \multicolumn{1}{c|}{0.8097}          & 0.4232             & \multicolumn{1}{c|}{0.8361}          & 0.4433             \\ \hline
Ours                      & \multicolumn{1}{c|}{0.7131} & \textbf{0.3102}    & \multicolumn{1}{c|}{\textbf{0.7843}} & \textbf{0.4211}    & \multicolumn{1}{c|}{\textbf{0.8036}} & \textbf{0.4443}    & \multicolumn{1}{c|}{\textbf{0.8418}} & \textbf{0.4816}    & \multicolumn{1}{c|}{\textbf{0.8507}} & \textbf{0.5071}    \\ \hline
\end{tabular}
\end{table*}

\begin{figure*}[t!]
\centering
\includegraphics[width=0.95\textwidth]{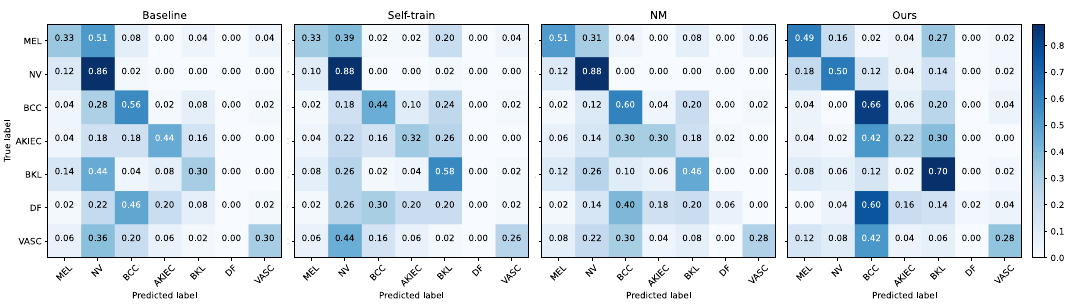}
\caption{Confusion matrices of different methods on the HAM10000 dataset when the number of labeled training samples is 800. \textbf{From left to right}: Baseline, Self-train, NM and Our method.}
\label{fig:pathdemo3}
\end{figure*}

\subsection{Results on HAM10000}
In an effort to verify the efficacy of our proposed method, we select four competing state-of-the-art semi-supervised image classification algorithms, including two for dermoscopic image classification (GLM~\cite{gyawali2020semi} and NM~\cite{wang2021neighbor}), one for generic image classification (Self-train~\cite{grandvalet2004semi}), and a popular and recent consistency-based algorithm~(FixMatch \cite{sohn2020fixmatch}). We also report a baseline that trains the model using only the labeled data. For a fair comparison, we re-run all the algorithms under varying levels of labeled images based on the settings described in the previous section. The evaluation metrics include average AUC and MCA, which can better represent the performance of the algorithm in the case of an unbalanced dataset. 

As shown in Table~\ref{tab1}, our method outperforms the state-of-the-art methods under different labeled data ratios. Specifically, we make 1.23$\%$, 2.79$\%$, 3.65$\%$, 4.16$\%$, 2.37$\%$ improvements on AUC and 1.21$\%$, 5.43$\%$, 4.89$\%$, 6.89$\%$, 7.73$\%$ improvements on MCA over the baseline, and compared favorably to competing methods~\cite{sohn2020fixmatch, grandvalet2004semi, gyawali2020semi, wang2021neighbor} under most settings. Notably, in some settings, FixMatch and GLM perform worse than the baseline. In contrast, improvements from our method are more stable across all different settings. Some example qualitative results are shown in Fig.~\ref{fig:qualitative}.

In order to better interpret the classification results on individual categories, we present the confusion matrix on HAM10000 with $800$ labeled images in Fig.~\ref{fig:pathdemo3}, comparing our method against the Baseline, Self-train, and NM. Although the other methods provides a higher accuracy on the majority class NV, there is a systematic bias towards majority classes. Our approach, on the other hand, significantly improves the model's ability in correctly classifying minority classes such as BCC and BKL. Overall, our method outperforms other methods in terms of AUC and MCA, as already demonstrated in Table~\ref{tab1}.

\subsection{Ablation Studies on HAM10000}

\subsubsection{Comparison with DA}
\begin{figure*}[t!]
\centering
\includegraphics[width=0.95\textwidth]{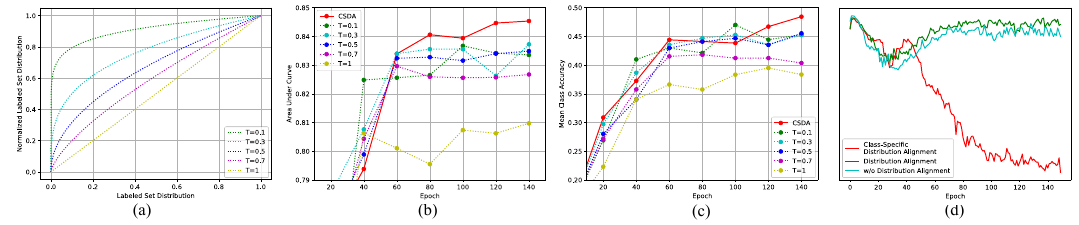}
\caption{Quantitative analysis of temperature scaling and class-wise distribution distance. \textbf{(a)} Function graph of different temperature values. \textbf{(b) and (c)} Influence of our adaptive temperature strategy~(shown as CSDA) \textit{vs.} constant temperature on AUC and MCA. \textbf{(d)} The Frobenius norm distance between the class-wise labeled and unlabeled distributions after alignment.}
\label{fig5}
\end{figure*}

In this paper, we proposed the Class-Specific Distribution Alignment (CSDA) for semi-supervised medical image classification. Therefore, one of the most important questions that naturally arises is if CSDA indeed provides performance improvements over conventional Distribution Alignment (DA). As shown in Table~\ref{tab2}, for the evaluation metric AUC, CSDA outperforms DA by 1.4$\%$ to 2.6$\%$ under various labeled data ratios, while for MCA the performance improvements are between 1.1$\%$ to 9.2$\%$. We note that these improvements we obtain are consistent across different settings, which suggest that CSDA is indeed superior to DA especially when labeled data is scarce and unbalanced. In order to illustrate the role of CSDA in a more intuitive manner, we calculate the Frobenius norm distance between the class-wise labeled and unlabeled distributions after alignment, as shown in Fig.~\ref{fig5} (d). It is clear that the class-agnostic distribution in DA is not sufficient to represent the changing distribution across different classes. In addition, using the Variable Condition Queue (VCQ) alone also improves performance over DA, indicating that VCQ can obtain samples with appropriate proportion to better train models when the class distribution is unbalanced. We also present an upper-bound of semi-supervised learning methods by using the ground-truth labels of unlabeled data to train the model, which are typically unavailable during training. The large performance gap between our method and the upper-bound suggests that the problem itself is challenging, and there is still a large room of improvement for semi-supervised learning methods.

\begin{table*}[t!]
\centering
\caption{Ablation analysis on the HAM10000 dataset. The first row indicates the number of labeled training samples.}
\label{tab2}
\setlength{\tabcolsep}{1.8mm}
\begin{tabular}{l|cc|cc|cc|cc|cc}
\hline
\multirow{2}{*}{Method} & \multicolumn{2}{c|}{175}                               & \multicolumn{2}{c|}{350}                               & \multicolumn{2}{c|}{600}                               & \multicolumn{2}{c|}{800}                               & \multicolumn{2}{c}{1200}                               \\ \cline{2-11} 
                        & \multicolumn{1}{c|}{AUC}             & MCA             & \multicolumn{1}{c|}{AUC}             & MCA             & \multicolumn{1}{c|}{AUC}             & MCA             & \multicolumn{1}{c|}{AUC}             & MCA             & \multicolumn{1}{c|}{AUC}             & MCA             \\ \hline
Upper-bound             & \multicolumn{1}{c|}{0.8654}          & 0.5156          & \multicolumn{1}{c|}{0.8654}          & 0.5156          & \multicolumn{1}{c|}{0.8654}          & 0.5156          & \multicolumn{1}{c|}{0.8654}          & 0.5156          & \multicolumn{1}{c|}{0.8654}          & 0.5156          \\ 
Baseline                & \multicolumn{1}{c|}{0.6880}          & 0.2865          & \multicolumn{1}{c|}{0.7545}          & 0.3980          & \multicolumn{1}{c|}{0.7689}          & 0.3665          & \multicolumn{1}{c|}{0.8098}          & 0.4187          & \multicolumn{1}{c|}{0.8311}          & 0.4269          \\ 
Ours(w/o CSDA, w/DA)     & \multicolumn{1}{c|}{0.6987}          & 0.2895          & \multicolumn{1}{c|}{0.7605}          & 0.4037          & \multicolumn{1}{c|}{0.7866}          & 0.3668          & \multicolumn{1}{c|}{0.8153}          & 0.3897          & \multicolumn{1}{c|}{0.8365}          & 0.4610          \\ 
Ours(w/o CSDA, w/VCQ)           & \multicolumn{1}{c|}{0.7077}          & 0.2896          & \multicolumn{1}{c|}{0.7743}          & 0.4093          & \multicolumn{1}{c|}{0.8035}          & 0.4122          & \multicolumn{1}{c|}{0.8199}          & 0.4269          & \multicolumn{1}{c|}{0.8375}          & 0.4554          \\ \hline
Ours           & \multicolumn{1}{c|}{\textbf{0.7131}} & \textbf{0.3102} & \multicolumn{1}{c|}{\textbf{0.7843}} & \textbf{0.4211} & \multicolumn{1}{c|}{\textbf{0.8036}} & \textbf{0.4443} & \multicolumn{1}{c|}{\textbf{0.8418}} & \textbf{0.4816} & \multicolumn{1}{c|}{\textbf{0.8507}} & \textbf{0.5071} \\ \hline
\end{tabular}
\end{table*}

\subsubsection{Influence of Temperature Scaling}
In our method, we use a temperature parameter $T_i$ to scale the marginal distribution for the labeled data $\boldsymbol{ \lambda^x_i}$, i.e., $\text{Normalize}\Big(\big(\boldsymbol{\lambda_i^x}\big)^{T_i}\Big)$ is used as the labeled set distribution. In particular, we propose a class-adaptive temperature in Eqn.~\ref{eqn:temperature}. Here, we first show the influence of different temperatures on the distribution in Fig.~\ref{fig5} (a). We can see that smaller temperatures lead to more intense change of values. When $T_i=1$, the labeled distribution will not change. To demonstrate the superiority of class-adaptive temperatures, we present in Fig.~\ref{fig5} (b) and (c) the effects of setting the temperature as a constant and our adaptive temperature strategy (CSDA) on AUC and MCA. Firstly, we can observe that it is effective to scale the labeled distribution with temperature. For example, when $T_i=0.7$, the performance in terms of AUC is clearly better than $T_i=1$ by about 2$\%$. Fine-tuning $T_i$ will lead to better AUC, but the AUC may decrease slightly in the later stage of training, partially because the model has now been able to output relatively accurate unlabeled predictions. Still using a constant temperature at this time will have a negative effect, widening the gaps between the unlabeled distributions and the labeled distributions. In contrast, our adaptive temperature scaling strategy is not only able to obtain a higher AUC, but also maintain a stable performance during the entire training process.

\begin{table*}[t!]
\centering
\caption{Performance impact of the queue entry threshold $\delta$ on the HAM10000 dataset with 800 labeled training samples.}
\label{tab3}
\setlength{\tabcolsep}{1.8mm}
\begin{tabular}{l|c|c|c|c|c|c|c}
\hline
Metric & \multicolumn{1}{l|}{($\delta$=0.1)} & \multicolumn{1}{l|}{($\delta$=0.15)} & \multicolumn{1}{l|}{($\delta$=0.2)} & \multicolumn{1}{l|}{($\delta$=0.25)} & \multicolumn{1}{l|}{($\delta$=0.3)} & \multicolumn{1}{l|}{($\delta$=0.35)} & \multicolumn{1}{l}{($\delta$=0.4)}  \\ \hline
AUC    & 0.8253                              & 0.8139                               & 0.8274                              & \textbf{0.8418}                      & 0.8104                              & 0.8264                               & 0.8228                              \\ 
MCA    & 0.4471                              & 0.4416                               & 0.4841                              & \textbf{0.4816}                      & 0.4299                              & 0.4529                               & 0.4442                              \\ \hline
      & \multicolumn{1}{l|}{($\delta$=0.5)} & \multicolumn{1}{l|}{($\delta$=0.75)} & \multicolumn{1}{l|}{($\delta$=0.8)} & \multicolumn{1}{l|}{($\delta$=0.85)} & \multicolumn{1}{l|}{($\delta$=0.9)} & \multicolumn{1}{l|}{($\delta$=0.95)} & \multicolumn{1}{l}{($\delta$=0.99)} \\ \hline
AUC    & 0.8184                              & 0.8272                               & 0.8217                              & 0.8328                               & 0.8277                              & 0.8252                               & 0.8285                              \\ 
MCA    & 0.4614                              & 0.4443                               & 0.4471                              & 0.4501                               & 0.4672                              & 0.4407                               & 0.4586                              \\ \hline
\end{tabular}
\end{table*}

\subsubsection{Analysis of Queue Entry Threshold}
In Table~\ref{tab3}, we present results obtained with varying queue entry threshold $\delta$ in Eqn.~\ref{condition_threshold} when we use $800$ labeled images on the HAM10000 dataset. When $\delta=0.25$, AUC reaches the highest value of 84.18$\%$, while for MCA the highest is at $\delta=0.2$, which is 48.41$\%$. This observation confirms that it is necessary to set a threshold for the minimum prediction probability to enqueue. In particular, the threshold should be relatively low to allow a larger number of samples to enter our VCQ, so that the model could learn from richer semantic features. In the case of skin lesion classification, we note that the pathological regions of some specific skin diseases are relatively similar, therefore different types of skin disease images will improve each other to obtain appearance features of higher quality.

\subsubsection{Impact of Queue Length}

The different ratio of unlabeled data will affect the recognition performance of semi-supervised classification tasks~\cite{zhang2021flexmatch}. Therefore, we now move on to explore the performance impact from the changing ratio of unlabeled data in each minibatch. We do so by changing the maximum queue length $L$. Due to the fact that we are using a fixed number of labeled images (i.e., 128) in each minibatch, increasing the maximum queue length $L$ will result in a higher relative proportion of unlabeled data. As shown in Table~\ref{tab4}, the maximum queue length is approximately directly proportional to AUC and MCA, which means that the increase of unlabeled data ratio is helpful for our semi-supervised learning task. However, when $L$ is 256, AUC and MCA will decrease slightly, probably because the data distribution deviates from the ground truth distribution with this specific queue length setting that is not conducive to the learning of the model.

\begin{table}[t!]
\centering
\caption{Queue Length $L$ analysis on the HAM10000 dataset with 800 labeled training samples.}
\label{tab4}
\setlength{\tabcolsep}{1.2mm}
\begin{tabular}{c|c|c|c|c|c}
\hline
\multirow{2}{*}{Method} & \multicolumn{2}{c|}{Number of samples}          & \multirow{2}{*}{\begin{tabular}[c]{@{}c@{}}Queue\\ Length $L$\end{tabular}} & \multirow{2}{*}{AUC} & \multirow{2}{*}{MCA} \\ \cline{2-3}
                        & \multicolumn{1}{c|}{Labeled} & Unlabeled &                                                                         &                      &                      \\ \hline
Baseline                & \multicolumn{1}{c|}{800}     & 0         & n/a                                                      & 0.8098               & 0.4187               \\ \hline

\multirow{5}{*}{Ours} & \multirow{5}{*}{800} & \multirow{5}{*}{8615}  & 16                                                                      & 0.8226               & 0.4354               \\
& & & 64  & 0.8309 & 0.4585 \\
& & & 128 & 0.8341 & 0.4701 \\ 
& & & 256 & 0.8336 & 0.4509 \\
& & & 512 & \textbf{0.8418} & \textbf{0.4816} \\ \hline

\end{tabular}
\end{table}

\begin{table}[t!]
\centering
\caption{Resampling parameter $\gamma$ analysis on HAM10000 dataset with 800 labeled training samples.}
\setlength{\tabcolsep}{1mm}
\label{tab5}
\begin{tabular}{c|c|c|c|c|c}
\hline
\multirow{2}{*}{Method} & \multicolumn{2}{c|}{Number of samples}          & \multirow{2}{*}{$\gamma$} & \multirow{2}{*}{AUC} & \multirow{2}{*}{MCA} \\ \cline{2-3}
                        & \multicolumn{1}{c|}{Labeled} & Unlabeled &                           &                      &                      \\ \hline
Baseline                & \multicolumn{1}{c|}{800}     & 0         & n/a        & 0.8098               & 0.4187               \\ \hline
\multirow{3}{*}{Ours} & \multirow{3}{*}{800} & \multirow{3}{*}{8615}      & 0                         & 0.8178               & 0.4357               \\ 
& & & 1                         & \textbf{0.8418}      & \textbf{0.4816}      \\
& & & 0.5                       & 0.8227               & 0.4475               \\ \hline
\end{tabular}
\end{table}

\subsubsection{Effect of the Resampling Parameter}
Since the resampling parameter $\gamma$ is a key factor affecting the class distribution in our VCQ, we also empirically compare different choices of this parameter here. As shown in Table~\ref{tab5}, AUC obtained by instance-based sampling ($\gamma=1$) is 2.4$\%$ and 1.91$\%$ higher than class-based sampling ($\gamma=0$) and square-root sampling ($\gamma=0.5$) respectively, and the corresponding MCA is 4.59$\%$ and 3.41$\%$ higher than the other two sampling methods. One of the key advantages of instance-based sampling is that it selects samples from a particular class proportionate to the mean class confidence in the training set. In this way, we can make the class distribution in VCQ closer to the underlying ground-truth class distribution, so as to minimize the class distribution mismatch between the labeled set and the the unlabeled set.

\subsection{Results on CheXpert}
\begin{figure}[t!]
\centering
\includegraphics[width=0.60\textwidth]{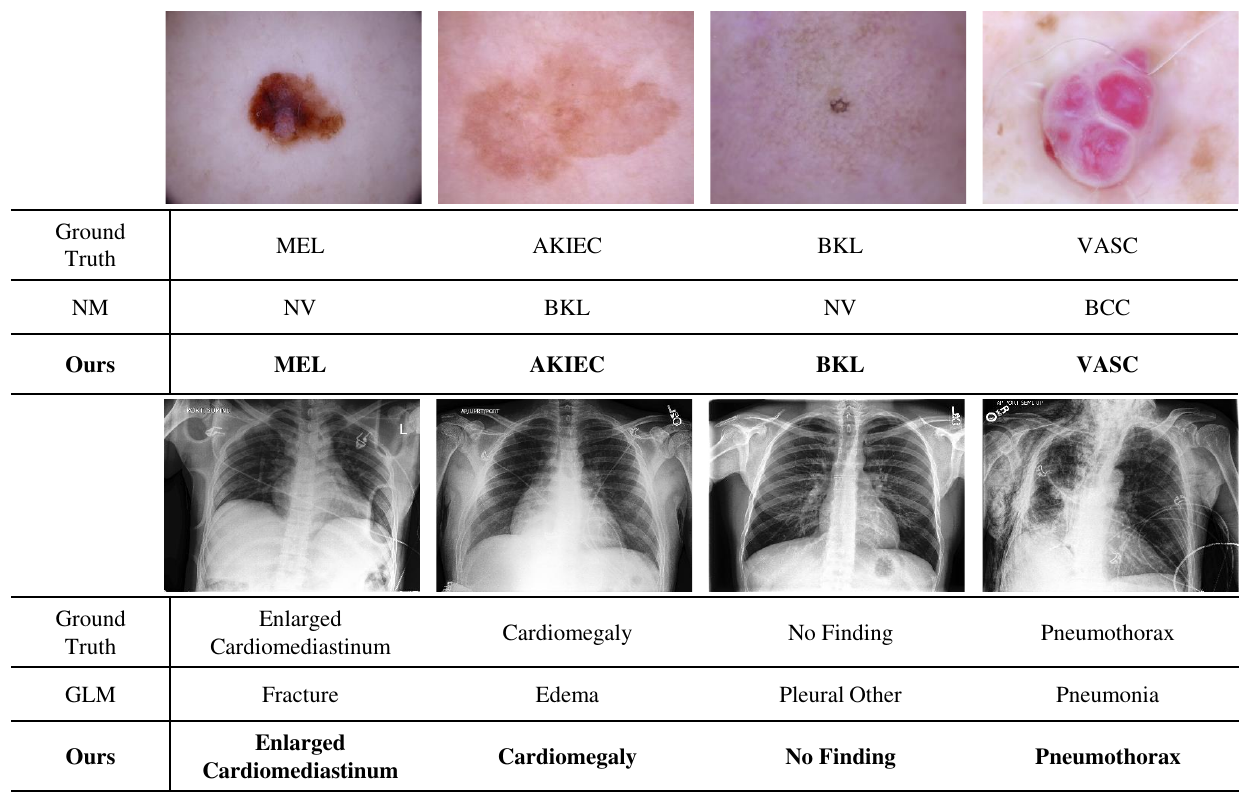}
\caption{Example qualitative classification results on HAM10000 and CheXpert. The ground truth labels are shown in the first row below the image, the pseudo-labels produced by NM~\cite{wang2021neighbor} are listed in the second row, the pseudo-labels produced by CSDA are shown in the last row (in bold).}
\label{fig:qualitative}
\end{figure}

In this section, we present further experimental evaluation results on the public chest disease classification dataset CheXpert. As shown in Table~\ref{tab6}, our method consistently surpasses two of the state-of-the-art methods, LSSE and BiS, with different numbers of labeled images. The mean AUC improvements over LSSE are 2.99$\%$, 2.33$\%$, 2.1$\%$, 1.25$\%$ and 0.97$\%$, respectively. On the other hand, unlike on HAM10000, GLM provides slightly better results than ours. This is probably due to that fact that GLM uses much stronger data augmentation via mixing at both the input and the feature space, which is orthogonal to the contributions made by our method. We note that GLM is not as effective on the HAM10000 dataset, suggesting that combining our method with GLM may produce even better results, which is out of the scope of this paper. We present some example qualitative results in Fig.~\ref{fig:qualitative}.

\begin{table}[t!]
\centering
\caption{Performance (AUC) on the CheXpert dataset.}
\label{tab6}
\setlength{\tabcolsep}{1.8mm}
\begin{tabular}{l|c|c|c|c|c}
\hline
\multirow{2}{*}{Method} & \multicolumn{5}{c}{Number of labeled training samples}                                                                                          \\ \cline{2-6} 
                        & \multicolumn{1}{c|}{100} & \multicolumn{1}{c|}{200} & \multicolumn{1}{c|}{300} & \multicolumn{1}{c|}{400} & 500 \\ \hline
Baseline                                     & 0.5576 & 0.6166 & 0.6208 & 0.6343 & 0.6353 \\ 
LSSE\cite{gyawali2019semi}                                         & 0.6200 & 0.6386 & 0.6484 & 0.6637 & 0.6697 \\ 
GLM\cite{gyawali2020semi}                                          & \textbf{0.6512} & \textbf{0.6641} & \textbf{0.6739} & \textbf{0.6796} & \textbf{0.6847} \\ 
BiS\cite{2021Rethinking}                                          & 0.5990 & 0.6138 & 0.6222 & 0.6355 & 0.6565 \\ \hline
Ours                                         & 0.6499 & 0.6619 & 0.6694 & 0.6762 & 0.6792 \\ \hline
\end{tabular}
\end{table}

\subsection{Results on Kvasir}
Table~\ref{tab7} shows the experimental results of our method and other competing methods on the Kvasir dataset. It can be seen that our method outperforms all other methods except BiS. However, although the data re-sampling strategy in BiS perform favorably on Kvasir, our method still outperforms BiS on both the HAM10000 dataset and the CheXpert dataset.
\begin{table*}[t!]
\centering
\caption{Performance on the Kvasir dataset. The first row indicates the number of labeled training samples.}
\label{tab7}
\setlength{\tabcolsep}{3.8mm}
\begin{tabular}{l|cc|cc|cc|cc}
\hline
\multicolumn{1}{c|}{\multirow{2}{*}{Method}} & \multicolumn{2}{c|}{1653}                              & \multicolumn{2}{c|}{3306}                              & \multicolumn{2}{c|}{6612}                              & \multicolumn{2}{c}{9918}                               \\ \cline{2-9} 
\multicolumn{1}{c|}{}                        & \multicolumn{1}{c|}{AUC}             & MCA             & \multicolumn{1}{c|}{AUC}             & MCA             & \multicolumn{1}{c|}{AUC}             & MCA             & \multicolumn{1}{c|}{AUC}             & MCA             \\ \hline
Baseline                                     & \multicolumn{1}{c|}{0.8136}          & 0.1644          & \multicolumn{1}{c|}{0.8304}          & 0.1708          & \multicolumn{1}{c|}{0.8359}          & 0.1812          & \multicolumn{1}{c|}{0.8496}          & 0.1893          \\
FixMatch~\cite{sohn2020fixmatch}                                     & \multicolumn{1}{c|}{0.8517}          & 0.1682          & \multicolumn{1}{c|}{0.8710}          & 0.2072          & \multicolumn{1}{c|}{0.8632}          & 0.2099          & \multicolumn{1}{c|}{0.8652}          & 0.2021          \\
Self-train~\cite{grandvalet2004semi}                                   & \multicolumn{1}{c|}{0.8180}          & 0.1700          & \multicolumn{1}{c|}{0.8403}          & 0.1841          & \multicolumn{1}{c|}{0.8663}          & 0.2024          & \multicolumn{1}{c|}{0.8572}          & 0.1954          \\
GLM~\cite{gyawali2020semi}                                          & \multicolumn{1}{c|}{0.8753}          & 0.2174          & \multicolumn{1}{c|}{0.8845}          & 0.2139          & \multicolumn{1}{c|}{0.8925}          & 0.2292          & \multicolumn{1}{c|}{0.8899}          & 0.2257          \\
NM~\cite{wang2021neighbor}                                           & \multicolumn{1}{c|}{0.8518}          & 0.1805          & \multicolumn{1}{c|}{0.8610}          & 0.2144          & \multicolumn{1}{c|}{0.8828}          & 0.2460          & \multicolumn{1}{c|}{0.8959}          & 0.2382          \\
BiS~\cite{2021Rethinking}                                          & \multicolumn{1}{c|}{\textbf{0.9296}} & \textbf{0.4776} & \multicolumn{1}{c|}{\textbf{0.9600}} & \textbf{0.5662} & \multicolumn{1}{c|}{\textbf{0.9787}} & \textbf{0.7957} & \multicolumn{1}{c|}{\textbf{0.9850}} & \textbf{0.8411} \\ \hline
Ours                                         & \multicolumn{1}{c|}{0.8861}          & 0.3529          & \multicolumn{1}{c|}{0.8991}          & 0.3905          & \multicolumn{1}{c|}{0.9353}          & 0.3794          & \multicolumn{1}{c|}{0.9439}          & 0.3898          \\ \hline
\end{tabular}
\end{table*}

\section{Discussions}
Our work introduces CSDA that provides new insights into distribution alignment by presenting a general form of mathematical formulation inspired by a change of basis in the vector space spanned by marginal predictions. In addition, our Variable Condition Queue module is able to maintain a proportionately balanced number of unlabeled samples for each class. In this section, we briefly discuss the main strengths and limitations of the proposed method.

\noindent \textbf{Strengths.} A key advantage of our method is that it is suitable for learning from imbalanced datasets. Specifically, the proposed CSDA and VCQ are able to produce a more balanced pseudo-label class distribution, i.e., the number of unlabeled samples that belong to minority classes will increase, alleviating the negative impact from highly imbalanced medical image datasets. In addition, we design our method to be a simple yet flexible module that could be inserted into any existing semi-supervised learning methods based on self-training, in order to refine the pseudo-labels.

\noindent \textbf{Assumptions.} In order for the proposed method to be effective, we conjecture that the following basic assumptions should be satisfied: the labeled dataset, despite its limited size, should be representative of the overall data distribution; the unlabeled data should contain useful information about the data distribution; and the decision boundary needs to be adequately supported by the labeled data.

\noindent \textbf{Limitations.} We note that our method does not provide an end-to-end trainable solution to semi-supervised medical image classification. In fact, it is possible to integrate the distribution transformation as a trainable module that is part of the convolutional neural network. In addition, as shown in Eqn.~\ref{eq5}, we have derived a rather general formulation for distribution alignment but we limit ourselves in this work to a special case that eliminates cross-class correlations between labeled and unlabeled marginal distributions. Finally, even though our method is able to address the negative impact from data imbalance, it is still difficult to perform satisfactorily on some classes with very little data, or when the data distribution in the training set differs significantly from that of the test set.

\noindent \textbf{Implications for Medicine Biology.} The potential significance of our method in medicine and biology lies in its ability to improve classification accuracy and reduce the need for manual annotation, thereby enabling more efficient and accurate analysis of large-scale medical image datasets. Besides, our method can be used to evaluate the efficacy and toxicity of potential drug candidates in drug discovery. In cellular and molecular biology, our method can be used to analyze fluorescence microscopy images of cells and tissues.

\section{Conclusions}
\label{sec:conclusion}
In this work, we address the semi-supervised medical image classification problem based on the self-training framework. We provide a novel perspective into distribution alignment by considering the process as a change of basis in the vector space spanned by marginal predictions, and propose Class-Specific Distribution Alignment to capture the correlation between marginal predictions across different categories, in order to obtain pseudo-labels for unlabeled data with better quality. In addition, we present a Variable Condition Queue to store a proportionately balanced number of unlabeled samples for each class. Our method is particularly suitable for challenging learning scenarios when the class distribution is highly imbalanced. Experiments on two public datasets demonstrate the superiority of our method in semi-supervised medical image classification. We hope our method could provide new insights into semi-supervised learning methods that involves a distribution alignment component, and also more broadly to the self-training framework when dealing with imbalanced datasets. Future research may consider incorporating clinical knowledge and interpretability of the prediction results into semi-supervised learning for medical image analysis.

\vspace{5mm}
\noindent \textbf{Acknowledgments.} Tao Wang and Zuoyong Li are supported by the National Natural Science Foundation of China (61972187, 61703195) and Fujian Natural
Science Foundation (2022J011112, 2020J02024). Tao Wang is also supported by the Research Project of Fashu Foundation (MFK23001).

%%%%%%%%%%%%%%%%%%%%%%%%%%%%%%%%%%%%%%%%%%
\bibliographystyle{ieee}
\bibliography{references}

\end{document}